# Detection and Localization of Multiple Image Splicing using MobileNet V1

**KALYANI KADAM[1], SWATI AHIRRAO[1], KETAN KOTECHA[2], SAYAN SAHU[3]**
[1, 3]Symbiosis Institute of Technology, Symbiosis International (Deemed University), Pune, Maharashtra 412115, India
[2] Symbiosis Centre for Applied Artificial Intelligence, Symbiosis International (Deemed University), Pune, Maharashtra 412115, India

**ABSTRACT** In modern society, digital images have become a prominent source of information and medium of communication. The easy availability of image-altering softwares have greatly reduced the expenses and expertise required to exploit visual tampering. Images can, however, be simply altered using these freely available image editing softwares. Two or more images are combined to generate a new image that can transmit information across social media platforms to influence the people in the society. This information may have both positive and negative consequences. Hence there is a need to develop a technique that will detect and locates a multiple image splicing forgery in an image. This research work proposes multiple image splicing forgery detection using Mask R-CNN, with a backbone as a MobileNet V1. It also calculates the percentage score of a forged region of multiple spliced images. The comparative analysis of the proposed work with the variants of ResNet is performed. The proposed model is trained and tested using our MISD dataset, and it is observed that the proposed model outperforms the variants of ResNet models (ResNet 51,101 and 151).

**Keywords** image forgery, multiple image splicing forgery, deep learning, MobileNet V1, Mask R-CNN

## I. INTRODUCTION

The human brain has an exceptional capacity for processing visual information. Most people respond to images more quickly than they do to texts. An image is worth a thousand words. Images are used in almost every area for communication, such as social media, news channels, military, court, insurance, education sector, entertainment business, health sector, and many more. With the development in image editing software tools and technologies available on portable devices such as smartphones and laptops, it is now possible to easily manipulate images for various purposes. These forged images may have a significant impact on society and can influence the views of people.

These days, social media campaigning has become a new trend in elections all around the world. On a more positive side, digital visuals are extensively employed to raise election awareness. At the same time, forged images with misinformation have been seen being distributed across social media to influence the public. According to a study [3], roughly 13.1% of Whatsapp posts were fraudulent during the last Brazilian presidential election. Furthermore, several fraudulent images containing misinformation regarding the COVID-19 pandemic recently went viral on social media platforms [4].

Several ways are available for forging the image such as image splicing and copy move. Image splicing [1-2] merges two images to create a spliced image. Copy move uses a single image; in this, one object is copied into the same image. As a result, forgery detection techniques must be developed to ensure the authenticity of such images. Many researchers have proposed passive and active forgery detection techniques to authenticate digital images in recent years such as [1-2]. The active forgery detection technique detects forgery in an image with the help of statistical information of an image. On the other hand, the passive method doesn't require such information to detect forgeries. Instead, they detect the forgery using the features of an image.

In CV, the techniques used in earlier days for image splicing forgery detection rely on the traditional features extraction methods. These features are primarily selected to focus on specific image properties and are generated manually. Due to this, these feature extraction methods are also referred to as handcrafted feature extraction methods. Some of the prominent handcrafted features used in image splicing forgery detection are the DWT [5], LBP [6], CT [7], HHT [8], and DCT [6], Bi-coherence, camera response operation, invariant image moments. The limitation of handcrafted features is that they are not robust and computationally heavy due to high dimensions.

The DL techniques show extraordinary performance in various areas such as image processing, digital image forensics [9,10], fraud detection, self-driving cars, virtual assistance, and face recognition system. Recent developments ([11], [12], [13], [14], [15]) have focused on DL-based image splicing detection, as compared to hand-crafted feature-based image splicing detection techniques. DL-based techniques can learn more generic features from the input image in general. As a result, in recent years, DL-based image splicing forgery detection algorithms have grown in popularity.

There are various region based networks such as R-CNN [16], Fast R-CNN [17], Faster R-CNN [18], and Mask R-CNN [19] are available for object detection and segmentation. The R-CNN[16] excerpts many RPs from the input image, then utilizes a CNN on each RP to excerpt its features, which are then used to predict the RP's class and bounding box. In R-CNN near around 2000 image proposals are sent to CNN. As a result, utilizing R-CNN to train and test the image is computationally expensive. To address this issue, the Fast R-CNN architecture was created, which takes the entire image as input. It also introduces the area of interest pooling layer, which allows features of the same shape to be retrieved for different-shaped ROI. To improve object detection accuracy, the fast R-CNN[17] model must generate a large number of region recommendations in selective search. The Faster R-CNN [18] replaces selective search with RPN to reduce region proposals without compromising accuracy. The Mask R-CNN [19] is the improved version of Faster R-CNN. It provides a class and bounding box for each ROI and it also provides the mask, i.e., the pixel-wise position of the object using FCN.

Various CNN networks have been introduced in the computer vision field, including AlexNet [20], which won the ILSVRC in the year 2012, increasing classification accuracy by 10% above typical machine learning algorithms. The University of Oxford's Visual Geometry Group suggested VGGNet [21] in 2014, and GoogLeNet [22], and ResNet [23] in 2015. To obtain increased accuracy, several CNN networks in the CV listed above are growing increasingly complicated. The depth and parameters of the DL networks listed above growing exponentially, making them more reliant on computationally efficient graphical processing units (GPUs) [24]. To overcome the limitations of previous research, this research work proposes a MobileNet V1-based lightweight DL classification network [25]. This network is based on the DCL [25][26], which reduces convolution processing complexity and network parameter values, resulting in a lightweight network. The research on image splicing forgery detection faces challenges below :

- Lack of publicly accessible standard and custom datasets for detection of Multiple Image Splicing forgeries.
- Lack of forgery detection techniques for the detection of multiple image splicing forgeries.
- Lack of lightweight models which estimate the percentage score of the forged region of a multiple spliced image

**Contributions**:
- Detection, localization, and identification of passive forgeries like multiple image splicing using Mask R-CNN with pretrained backbone networks such as ResNet 51, ResNet 101, ResNet 151, and MobileNet V1.
- Evaluation of multiple image splicing forgery detection on our Multiple Image Splicing Dataset (MISD) using pretrained networks such as ResNet 51, ResNet 101, ResNet 151 with MobileNet V1.
- Comparative analysis of Mobilenet V1 with variants of ResNet 51, ResNet 101, ResNet 151.
- To find out the percentage score for a forged region of a multiple spliced image.

This research paper structure is: Section 1 presents an introduction, sections 2 covers related work, section 3 outlines the MISD (Multiple Image Splicing Dataset) information and creation process, section 4 represents proposed architecture for multiple image splicing, section 5 outlines experimental setup, section 6 shows MISD details, section 7 specifies results, section8 gives limitation of proposed research work, and section 9 presents the conclusion.

II. **RELATED** WORK

Existing work on image splicing forgery detection is explored with respect to the dataset and deep learning models. This section discusses the dataset employed by researchers for the detection of image splicing forgery. Table 1 shows a summary of image splicing datasets used in image splicing forgery detection. Figure 1,2,3 and 4 shows sample image from Columbia Color, CASIA 1.0, CASIA 2.0 and WildWeb dataset.

   *A. Datasets for Image Splicing*
   *i. Columbia Gray [27]*

This dataset contains 1845 image blocks, out of which 933 are AU, and 912 are SP. The AU and SP image blocks are having a size of 128 x 128 pixels. These image blocks are in BMP image format with simple cut-paste operation without any post-processing operations. In this, the cut-paste operation is performed using Adobe Photoshop [28]. In this dataset, the image blocks are grayscale, retrieved from 322 photos, 10 are captured using a camera by the authors, and 312 are taken from the CalPhotos dataset[29]. The limitation of this dataset is it provides only grayscale image block, not the color. It also does not provide, the ground truth masks for spliced image blocks.
   *ii. Columbia Color [30]*

Columbia color dataset addresses the shortcomings of the Columbia Gray dataset. This dataset has 363 images, 183 are AU, and 180 are SP. All color images in this dataset are in TIFF image format, with image dimension range varying from 757 × 568 to 1152 × 768 in pixels. In this dataset, the authentic images are captured using cameras such as *canong3*, *nikond70,canonxt*, and kodakdcs330. The spliced images are constructed from authentic images using Adobe Photoshop. The images in the authentic category have indoor and outdoor scene images that contain various objects such as keyboards, books, tables, etc. For this dataset, edge masks are provided, which represent the spliced objects boundaries.

### iii.    CASIA 1.0 [31]

The CASIA 1.0 dataset consists of 1725 images, 800 out of which are AU and 925 are SP. All images in this dataset are of JPG image format with a dimension of 384× 256. The SP images are constructed using Adobe Photoshop by performing copy and paste operations on AU images.

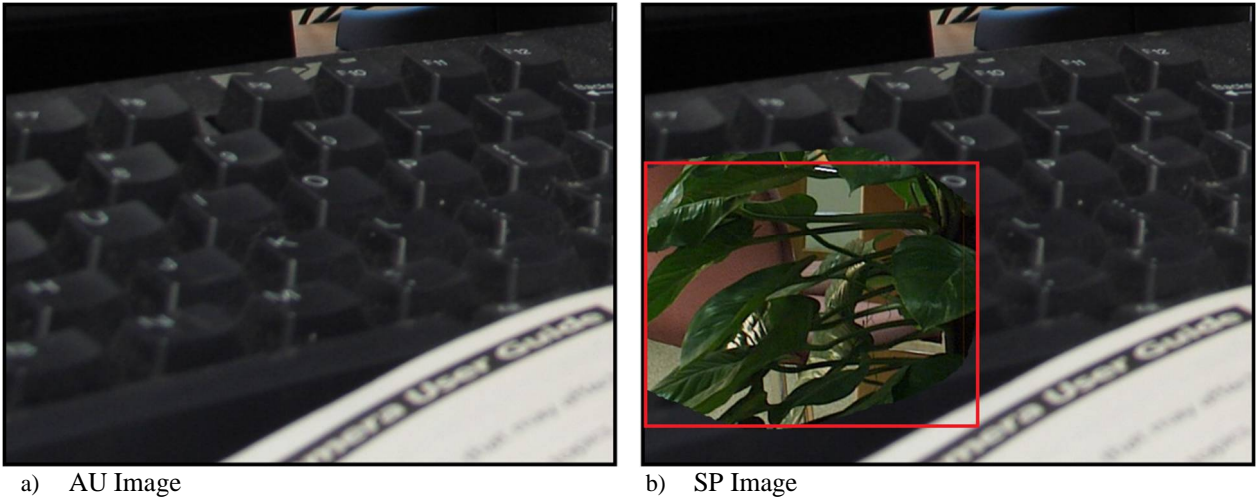

a)    AU Image          b)    SP Image

Figure 1.  **AU and SP Image from Columbia Color  dataset**

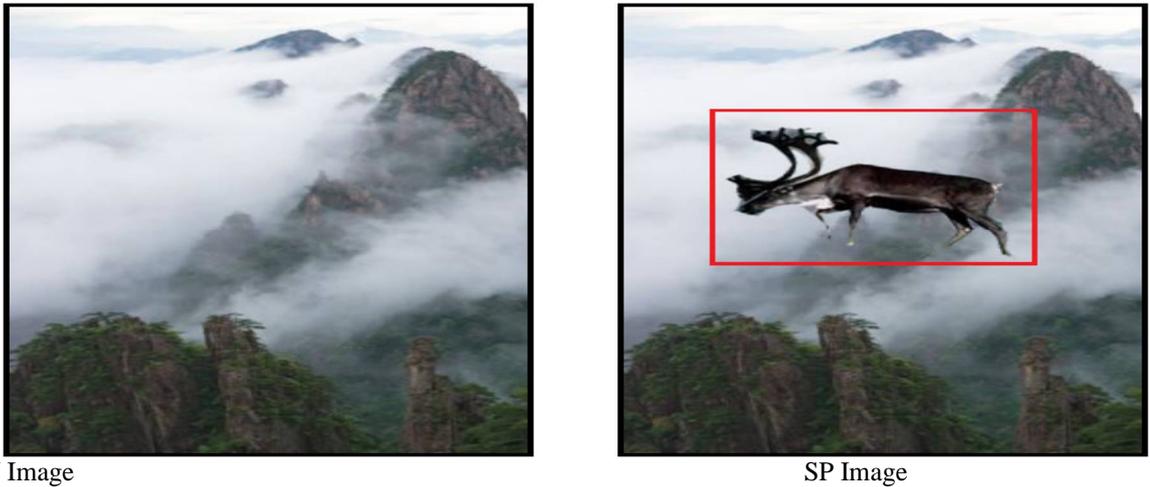

AU Image                                    SP Image

Figure 2.  **AU and SP Image from CASIA 1.0 dataset**

### iv.    CASIA 2.0 [31]

The CASIA 2.0 dataset contains a total of 12614 images, 7491 of which are AU images, and 5123 are forged. This dataset contains both copy move and image splicing images. Thus, there are 3274 images of copy move and 1849 images of image splicing. The images are in JPEG and TIFF image formats. For image splicing images, 753 out of the 1849 SP images are in TIFF format, while 1096 are in JPG image format. The dimension of images in pixels ranging from 320×240 to 800×600.

Casia 1.0 and Casia 2.0 are constructed using Adobe Photoshop CS3 version 10.0.1 on Windows XP. The images in these datasets are of various categories: *animal, architecture, art, indoor, nature, plant, scene, and texture*. But, both datasets do not provide ground truth mask information for copy move and spliced images.

    *v.*    **DSO-1 [32]**

This dataset contains 200 images, 100 AU, and 100 SP images, including indoor and outdoor images with image dimensions of 2048 × 1536 pixels. In this dataset, SP images are created by adding one or two people to the AU image. It applied post-processing operations to few SP images, such as color and brightness modification to create more realistic images.

    *vi.*    **DSI-1 [32]**

Carvalho et al. [32] constructed this dataset, and it contains a small set of popular image splicing categories acquired from the Internet. This dataset comprises 50 images, out of which 25 are AU, and 25 are SP of different dimensions.

    *vii.*    **WildWeb [33]**

This dataset's images are gathered via Internet sources. There are a total of 9666 spliced images created from 82 categories. The majority of the images in the dataset are in JPEG format, and the remaining are of type PNG, GIF, and TIFF. The images inside this dataset are difficult for splicing localization as they have gone through post-processing operations such as re-save and resample. In addition, this dataset includes a ground truth mask for spliced images. But, the dataset is not publicly accessible. However, it is available to the authors upon request for study purposes.

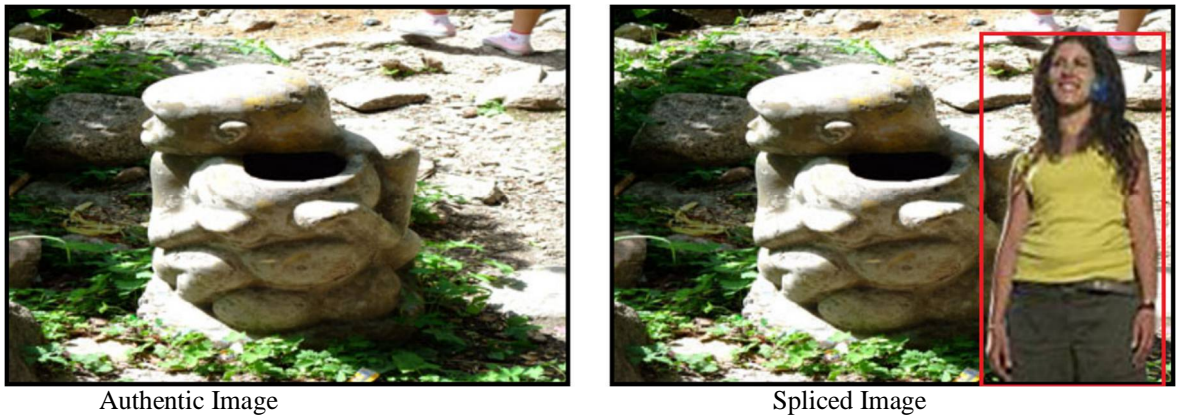

Authentic Image      Spliced Image

Figure 3 **AU and SP Image from CASIA 2.0 dataset**

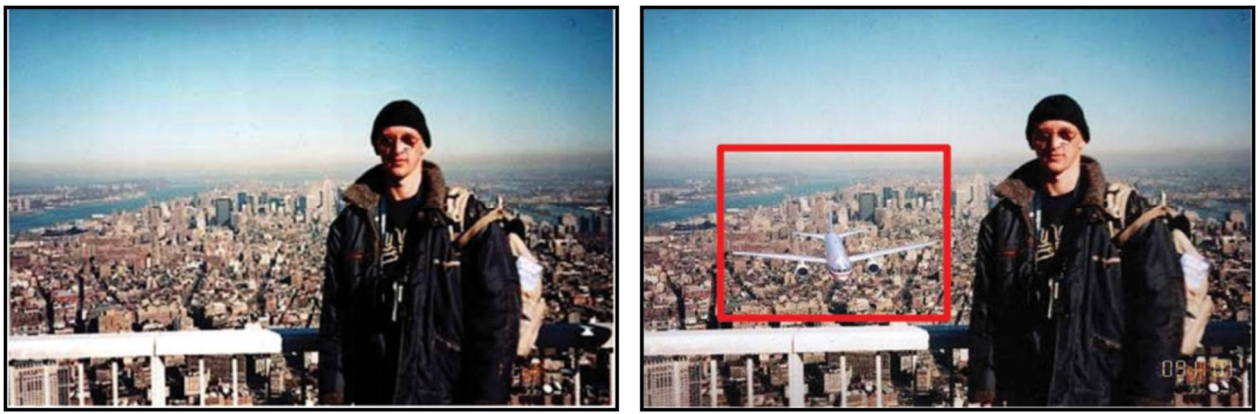

a)    AU Image      b)    SP Image

Figure 4. **AU and SP Image from WildWeb dataset**

**B.**   **Custom Dataset**

    **AbhAS [34]**

This dataset contains 93 images, out of which 45 are AU, and 48 are SP. The images in this dataset are of JPG image format with dimensions ranging from 278 × 181 to 3216 × 4288). In this dataset, 19 authentic images are taken from a single source camera, and the remaining 26 images are taken from the Internet. The spliced images are created using

Adobe Creative Cloud 2020 version with Photoshop. The ground truth masks are also available for these spliced images. A sample image from the AbhAS dataset is shown in Figure 5.

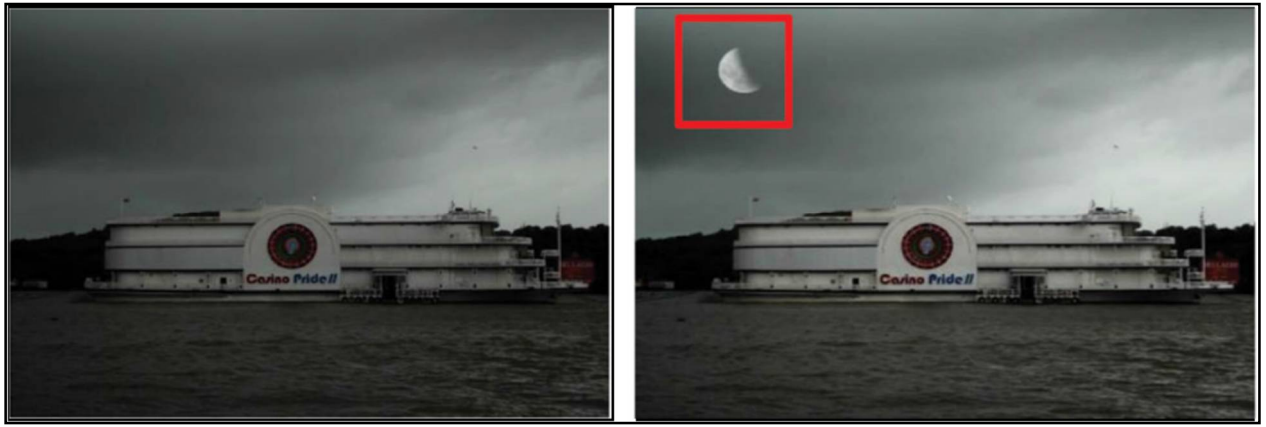

a) AU Image  b) SP Image

Figure 5. **AU and SP Image from AbhAS dataset**

TABLE 1  IMAGE SPLICING DATASETS

| Sr. No | Dataset Name | Type of Dataset | No of images | Size of image | Total No of Authentic images | Total No of Spliced images | Availability of ground truth mask |
|---|---|---|---|---|---|---|---|
| 1 | Columbia Gray [27] | Standard Datasets | 1845 image blocks | 128 x 128 | 933 | 912 | Not |
| 2 | Columbia Color [30] | | 363 | 757 × 568 to 1152 × 768 | 183 | 180 | Yes |
| 3 | CASIA 1.0 [31] | | 1725 | 384 × 256 | 800 | 925 | No |
| 4 | CASIA 2.0 [31] | | 12614 | 320 ×240 and 800 × 600 | 7491 | 1849 | No |
| 5 | DSO-1 [32] | | 200 | 2048 ×1536 and 1536 × 2048 | 100 | 100 | Yes |
| 6 | DSI-1 [32] | | 100 | Different sizes | 25 | 25 | Yes |
| 7 | WildWeb [33] | | 10666 | 122 × 120 to 2560 × 1600 | 100 | 9666 | Yes |
| 8 | AbhAS [34] | Custom Dataset | 93 | 278× 181 to 3216× 4288) | 45 | 48 | Yes |

**Challenges in the existing datasets**

I. **Standard Dataset**
- All the standard dataset contains splicing images which merge only two images for splicing operation.
- There are no standard datasets available for the detection of multiple image splicing forgery.
- There are no datasets available that are containing ground truth masks for multiple spliced objects.
- Some of the existing datasets have used only cut and paste operations for the creation of datasets.

II. **Custom Dataset**
- Very few images are available for image splicing. In addition, it does not contain multiple spliced images
- These datasets does not contain images from various categories.

C. **Deep Learning Models used for Image splicing**

This work [11] uses DL based approach for image splicing detection. In this, CNN is used to learn hierarchical features of the input image. The weights of this network's first layer are set to the value of the basic high-pass filter. It is used for the calculation of residual maps in the SRM. Next, the pretrained CNN is utilized as a patch descriptor to collect dense features from the test images. Then a feature fusion method is employed to get the final discriminative features for SVM classification. This research work [12] proposed a solution based on the ResNet-Conv deep neural network. Two variations of ResNet-Conv, ResNet-50, and ResNet-101, were utilized to build an initial feature map from RGB images. The authors also offered the *Mask-RCNN* for locating a forgery.

In this research work [35], two techniques such as combining resampling features and deep learning, are used to identify and locate image forgery. The first technique computes Radon transform of resampling features on overlapping image patches. A heatmap is then generated using deep learning classifiers and a Gaussian conditional random field model. Finally, a Random Walker segmentation approach is used to locate forged regions. In the second technique, resampling features obtained from overlapping image patches are passed to LSTM for classification and localization. Handcrafted features are frequently used to detect manipulated areas in a synthetic image to uncover and locate splicing forgeries. However, given a spliced image and no prior knowledge, determine which feature will be effective in exposing forgery.

Furthermore, a particular handcrafted feature can only deal with one type of splicing forgery. This research work proposes [36] a technique based on *deep neural networks* and *conditional random* to overcome this issue. Three distinct *FCNs* plus a *condition random field* are used to achieve this. Each *FCN* is trained to deal with image scales of varying sizes. *CRF* combines the detection findings from these neural networks in an adaptive way. The trained *FCNs–CRF* can subsequently be utilized to perform image authentication and forecast pixel-to-pixel forgery. Thus, *FCNs–CRF* model outperforms compared to existing techniques that rely on handcrafted features. This research work [37] proposes two *FCN*, to identify image splicing. The initial network is a single-task network that primarily learns the attributes of surface labels. The next network, on the other hand, is a two-path multi-task network. This two-path network primarily learns the spliced region's edge or boundary.

Recently researchers published a DL-based image splicing technique [38] that employs a convolutional neural network and a weight combination mechanism. Three distinct features were used in this method: YCbCr features, edge features, and PRNU features. These three features were combined, and their weight settings are automatically modified during the CNN training process, unlike the other approaches, until the best ratio is attained. This research work [39] uses ResNet 50 model and the 'Noiseprint' technique for image splicing forgery detection. Firstly the input image is preprocessed using the 'Noiseprint' technique to obtain the noise residual, suppressing the image content. Then ResNet-50 network is deployed for feature extraction. Finally, using SVM classifier, the collected features are classified as SP or AU.

Table 2. summarizes few DL techniques for image splicing forgery detection proposed in the existing literature.

TABLE 2 SUMMARY OF COMMONLY USED DL TECHNIQUES FOR IMAGE SPLICING DETECTION

| Paper | Technology | Dataset | Performance Metrics |
|---|---|---|---|
| [11] | CNN | CASIA V1.0[31],CASIA V2.0[31],Columbia Gray[12] | Accuracy on CASIA 1.0,CASIA 2.0 and Columbia Gray is 98.04%,97.83% and 96.38% respectively. |
| [12] | Mask R-CNN with backbone network as ResNet-conv | Computer-generated dataset where forged images have been generated using COCO and a set of objects with transparent backgrounds where 80,000 images are used for training and 40,000 for validation. The image size is $480 \times 640$ pixels. | AUC= 0.967 |
| [36] | CNN,LSTM | UCID[40] and RAISE [41] | Classification = 94.86% and AUC=0.9138 |
| [37] | FCN (Fully Convolution Network ), CRF (Condition Random Field) | CASIA V2.0[31] | F1-Score of the proposed method without compression = 0.4795 , JPEG Quality with 70 =0.4496, JPEG Quality with 50 = 0.4431 . F1-Score of the proposed method without noise = 0.4795 , SNR with 25 dB =0.4786, SNR with 20 dB =0.4811 , SNR with 15 dB =0.4719 |
| [38] | Fully Convolutional Networks (SFCN,MFCN, Edge-enhanced MFCN) | CASIA V1.0[31],CASIA V2.0[31],Columbia Gray[11], Nimble 2016 SCI, Carvalho [14] | F1- Scores on CASIA V1.0=0.5410 ,Columbia = 0.6117, Nimble 2016 SCI = 0.5707 , Carvalho = 0.4795 |
| [39] | Convolutional Neural Network (CNN) | CASIA V1.0[31],CASIA V2.0[31] | Detecttion Accuracy for CASIA 1.0 = 99.45 and for CASIA 2.0= 99.32 % |
| [40] | ResNet 50,SVM | Columbia Color [19] | Detection Accuracy = 97.24% |

III. MISD Dataset (Multiple Image Splicing Dataset)

This dataset contains 918 images, 618 of which are AU images, and 300 are multiple spliced images. The AU images are taken from Casia 1.0 [31] dataset. Multiple spliced images are created by performing various image editing operations with FIGMA software[42] on authentic images.

The images under AU are of nine categories: *animal, architecture, art, character, nature, indoor, scene, texture, and plant*. The ground truth masks are also available for these multiple spliced images. Table 3 gives the overall description of this dataset. Figure 6 shows the sample images for the Multiple Image Splicing dataset and figure 7 gives the different steps for the creation of this dataset.

TABLE 3 DESCRIPTION OF MULTIPLE IMAGE SPLICING DATASET.

|  | Number of images per category |  | Image size | Type of Image | Total number of images |
|---|---|---|---|---|---|
| **Authenticate Images** | Animal | 167 | 384 × 256 | JPG | 618 |
|  | Architecture | 35 |  |  |  |
|  | Art | 76 |  |  |  |
|  | Character | 124 |  |  |  |
|  | Indoor | 7 |  |  |  |
|  | Nature | 53 |  |  |  |
|  | Plant | 50 |  |  |  |
|  | Scene | 74 |  |  |  |
|  | Texture | 32 |  |  |  |
| **Multiple Spliced Images** | Images of all categories | 300 | 384 × 256 | JPG | 300 |

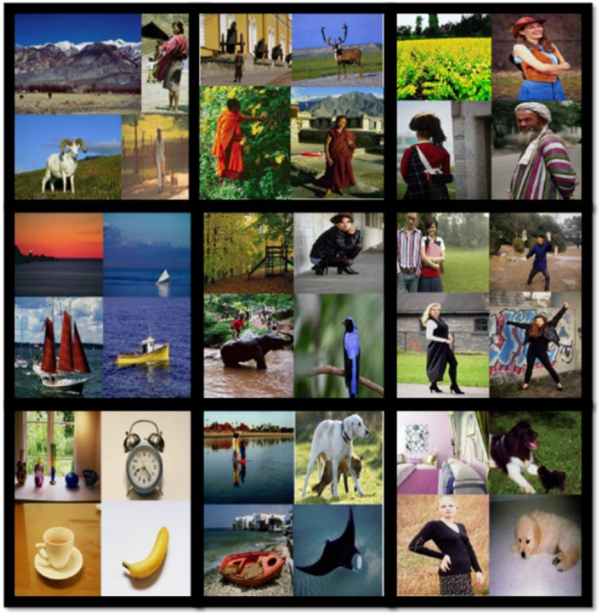
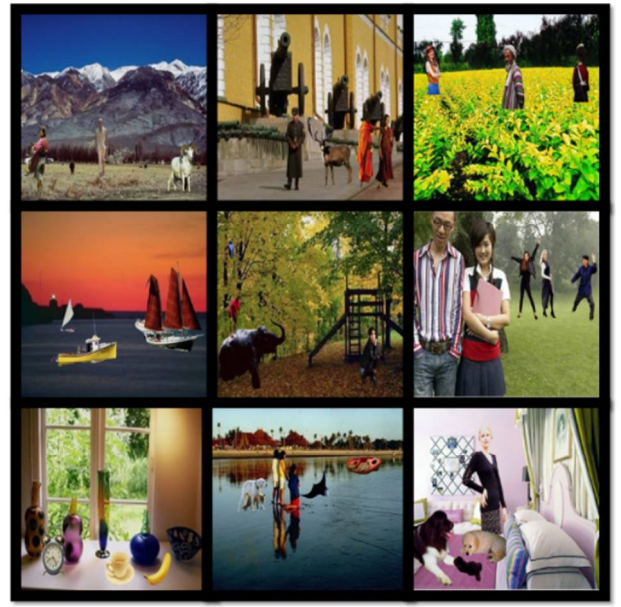

Figure 6. **Authentic and Spliced Image from MSID dataset**

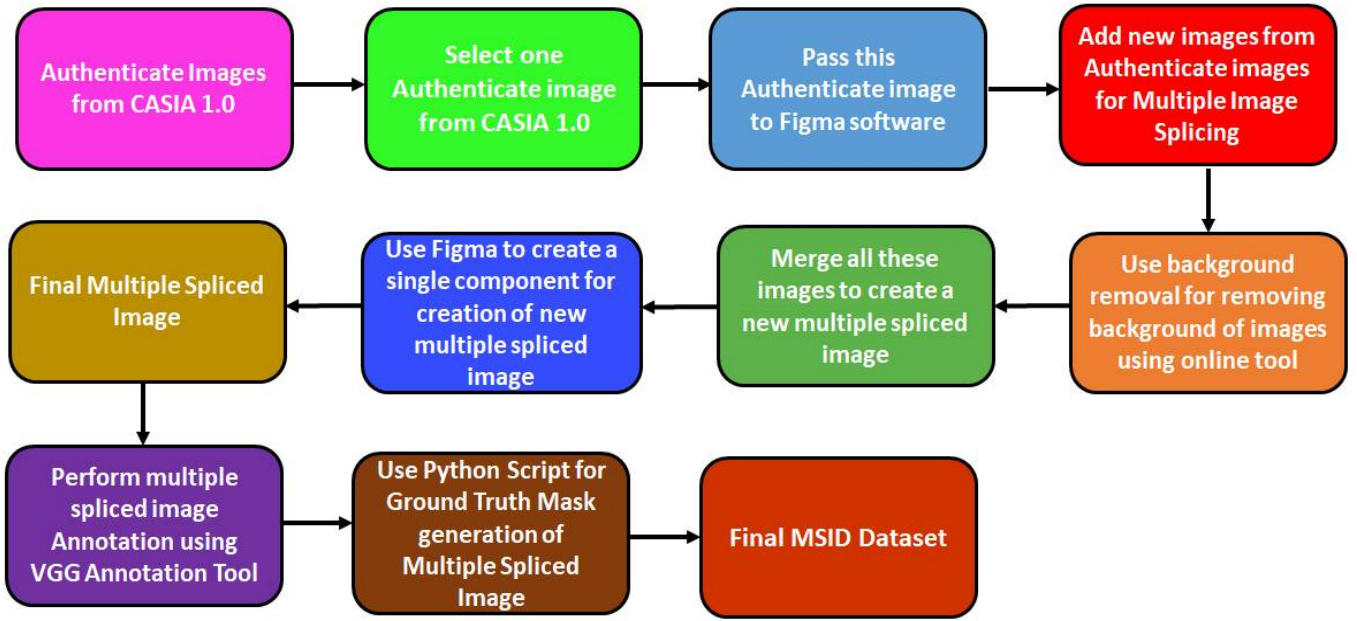

Figure 7. **MISD Dataset creation steps**

Following steps are followed for the construction of this dataset.
   a) Firstly, an image is uploaded into Figma software from authentic images. This image act as a source for the inclusion of various images.
   b) A background removal software, such as removing bg [43], cuts the objects from other Authenticate images. This software is used to remove the background of images. An image with a clear high contrast differentiation between the image's subject and background is preferred to achieve the best potential results. The generated image after the background removal is pasted on top of the base image using Figma software. The inserted objects are then subjected to different manipulation procedures such as transformation, rotation, brightness adjustment, and scaling to create the spliced images that appear more real and tougher to identify.
   c) Finally, all the added images/objects and the base image are selected, merged, and exported as a single image.
   d) The process is repeated with various authenticate images, and multiple images with backgrounds removed are added to the base image.
   e) Then these multiple spliced images are manually annotated using the VGG image annotation tool [ 44].
   f) Lastly, ground truth masks are generated for each multiple spliced image with a Python script which helps in identifying the spliced objects inside multiple spliced image.

IV. **Proposed Architecture for Multiple Splicing**

In this section, the proposed architecture shown in Figure 8 is used for multiple image splicing detection is explained. Our proposed system uses the Mask R-CNN framework, which is one of the futuristic object detection systems. The structure of Mask R-CNN is mainly made of three parts:
   1. The backbone convolutional neural network (CNN)
   2. Region Proposal Network (RPN)
   3. A fully convolutional network (FCN )

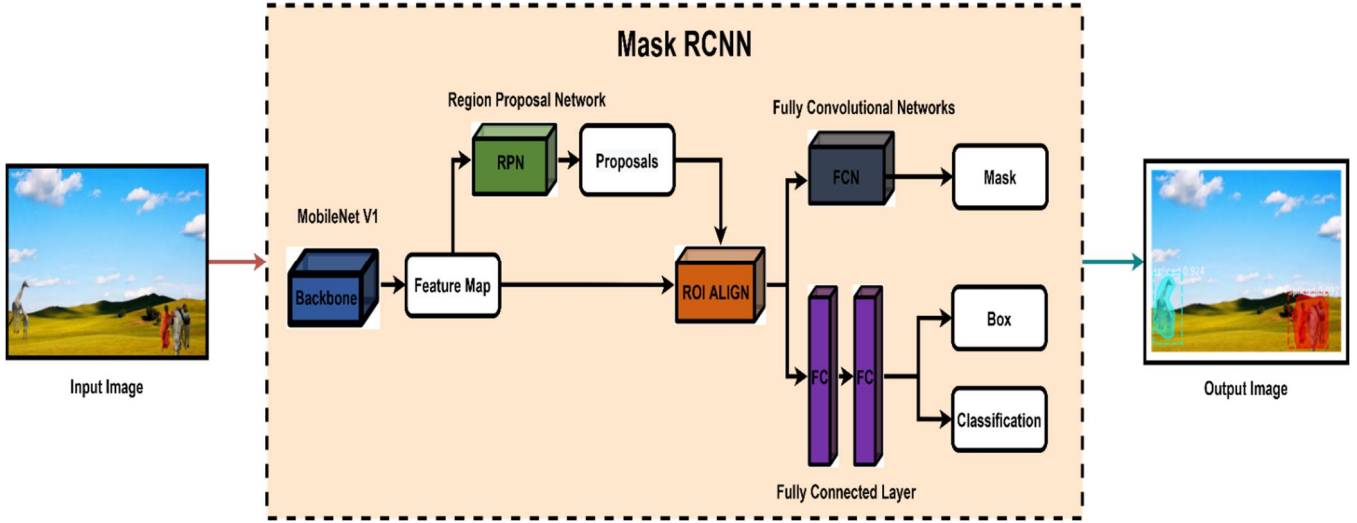

Figure 8. **Architecture for detection of Multiple Image Splicing**

       To detect multiple image splicing, the proposed network needs to extract and learn the features of a forged image. A feature can be any information relating to the edge, color consistency with the surrounding pixel, differences in brightness and contrast, object, or any region which makes the image a forged image. Two backbones are used as feature extractors in Mask R-CNN: a backbone convolutional neural network and a Feature Pyramid Network. In practice, the backbone convolutional neural network could be any one of the main models of deep neural networks, like ResNet, VGG, AlexNet, MobileNet, and GoogLeNet. In this research work, MobileNet V1 is used as a backbone network to extract the forged input image features. The goal was to reduce the network calculation parameters and speed up the detection process without reducing the accuracy of our proposed model. So, MobileNet V1 was used, as it uses depth-wise separable convolution to reduce the network capacity. According to the architecture of the MobileNet V1, MobileNet uses $3 \times 3$ DSL, which reduces the computation time by 8 to 9 times than using the standard convolutions.

       The architecture used in our experiment is shown in Table 4. Each row in the table represents a sequence of one or more identical layers, which repeats n times. For a layer in a particular sequence, the output channel is c. Apart from the first layer in each sequence whose stride is s, all other layers use a stride of 1. All depth-wise separable convolutions use $3 \times 3$ kernels. MobileNet uses five convolutional layers Conv1-5 in the RPN network to get the ROIs. The architecture of MobileNet shown in the table uses depth-wise separable convolution, which segregates a conventional convolution into a pointwise and depth-wise convolution. The depth-wise convolution has just one convolution kernel for a corresponding input channel. The pointwise convolution uses a $1 \times 1$ convolutional kernel to fuse the outputs from the depth-wise convolution linearly. There are no pooling layers present in between the depth-wise separable layers. Both convolutions are succeeded by a batch normalization layer and a Rectified linear unit (ReLU6) activation function, like ReLU, but it prevents the activations from becoming large. MobileNet uses two hyperparameters: depth multiplier and resolution multiplier. The depth multiplier is used to change the number of channels in each layer, and the resolution multiplier is used to control the resolution of the output image. These hyperparameters greatly optimize the computation speed and load.

TABLE 4 OVERALL ARCHITECTURE OF MOBILENET V1

| Layer Name | Input | Block Operator | c | n | s |
|---|---|---|---|---|---|
| Input Image | $224 \times 224 \times 3$ | Conv2d | 32 | 1 | 2 |
| Conv 1 | $112 \times 112 \times 32$ | DepthWiseConv | 64 | 1 | 1 |
|  | $112 \times 112 \times 64$ | DepthWiseConv | 128 | 1 | 2 |
| Conv 2 | $56 \times 56 \times 128$ | DepthWiseConv | 128 | 1 | 1 |
|  | $56 \times 56 \times 128$ | DepthWiseConv | 256 | 1 | 2 |
| Conv 3 | $28 \times 28 \times 256$ | DepthWiseConv | 256 | 1 | 1 |
|  | $28 \times 28 \times 256$ | DepthWiseConv | 512 | 1 | 2 |
| Conv 4 | $14 \times 14 \times 512$ | DepthWiseConv | 512 | 5 | 1 |
|  | $14 \times 14 \times 512$ | DepthWiseConv | 1024 | 1 | 2 |
| Conv 5 | $7 \times 7 \times 1024$ | DepthWiseConv | 1024 | 1 | 1 |
|  | $7 \times 7 \times 1024$ | - | - | - | - |

The ROI is generated directly on the feature map along with the Region Proposal Network (RPN). But in the ROI pooling, the coordinates in the feature map suffer from quantization, and the object location becomes misaligned. To avoid this and to accurately construct the ROI pool, Bilinear Interpolation called ROIAlign is used. The improved version of ROI pooling, i.e., Region of Interest Align (ROIAlign), uses Bilinear Interpolation instead of the rounding operation of the ROI pooling. It is used to get the pixel-level segmentation of the images. The ROIAlign is also used to maintain the exact spatial locations and return the feature map to a fixed size. Towards the end of the network, the FCN is used to pinpoint and classify the bounding boxes in the detection branch, and the mask is created on the image in the segmentation branch. In addition to this, to use the feature in a much better manner in each step, it also uses the Feature Pyramid Network (FPN) as a feature extractor to increase the accuracy and the speed. This acts as a replacement for feature extractors used in Faster R-CNN and constructs multi-scale feature maps to provide better information than the conventional feature pyramid.

In our proposed system, the Mask R-CNN for an input image uses the RPN network to propose the candidate ROI that may contain objects. In this layer, the SoftMax function is used to determine whether the anchors are positive or negative. To get accurate proposals, the anchors are modified with the help of bounding box regression. The backbone convolutional neural network MobileNet extracts the features from the image, and a pyramid feature map for the images is obtained. Since the standard operations for extracting the feature map from ROIs result in misalignments between the ROI and the extracted features due to quantization, this can greatly impact getting accurate masks and predictions. So, to correct the ROIs and negate the quantization operations, the ROIAlign layer is introduced. After getting the feature map of each ROI region, the classification and the bounding box of each ROI must be predicted. The FCN is applied to each ROI to predict the segmentation mask of the image tampering region in a pixel-to-pixel fashion. The FCN strategy arises from the traditional CNN network architecture but is also a little different from it. In CNNs, the convolutional layer is connected with numerous full connection layers. The FCN is quite similar to a CNN network, but the FCN network restores the output image size to that of the original image using deconvolution to up-sample the feature map. The Mask R-CNN loss for each proposal needs to be calculated; for this, we need to define a multitasking objective function to calculate the Mask R-CNN loss. This function includes $L_{cls}$ (classification loss), $L_{mask}$ (segmentation loss) and $L_{box}$ (bounding box location loss or the regression loss).

$$L = L_{cls} + L_{mask} + L_{box} \qquad (1)$$

The classification and the regression loss need to be calculated using the formula shown below:

$$L_{cls} + L_{box} = \frac{1}{N_{cls}} \sum_i L_{cls}(p_i, p_i^*) + \lambda \frac{1}{N_{reg}} \sum_i p_i^* L_{reg}(t_i, t_i^*) \qquad (2)$$

Here i is the index of an anchor, $p_i$ is the predicted probability of the anchor, $t_i$ are the four coordinate parameters of the box, $t_i^*$ is the four coordinate parameters of the ground truth box for the required positive anchor. $p_i^* = 1$ for a positive anchor, or else it is zero. The loss function has to be minimized to optimize the model

V. Experimental Setup

This section gives the experimental setup for multiple image splicing forgery detection. Tables 4,5, and 6 show the system specifications and parameters of the training environment. All experiments are conducted using *NVidia 1xTesla K80, compute 3.7, having 2496 CUDA cores with 12GB GDDR5 VRAM in google collaboratory*; the operating environment has *1xsingle core hyperthreaded Xeon Processors @2.3Ghz, i.e. (1 core, 2 threads) with 13 GB RAM. Tensorflow 1.8.0* is an open-source deep learning framework, and Python 3.7 is used as a programming language. COCO pretrained network [45] was used as the starting point to train the model. Table 4 shows few configuration parameters which were modified from the original Mask R-CNN. This experiment uses 734 images for training and 92 images for testing, and 92 images for validation purpose.

The training images were sized to maintain their aspect ratio. The mask size used is 28 × 28 pixels, and the size of the image is 512×512 pixels. This approach is different from the initial Mask R-CNN (He et al., 2020) approach, where the image resizing is done so that 800 pixels are the smallest size and 512 pixels are trimmed to the highest. Bbox selection is made by considering IOU, the ratio of expected bboxes to ground-truth boxes (GT boxes). Mask loss considers only positive ROI and is an intersection of ROI and its ground truth mask. Each mini-batch contains one image per GPU, where each image has an ROI of N samples and a 1:3 plus or minus ratio. The C4 backbone has a value of 64, while FPN has a value of 512. Images of batch size one were passed to the model on a single GPU unit. The model was trained for 360 iterations with an initial learning rate of 0.01, then modified to 0.003 at epoch 120 and 0.001 at epoch 240. Stochastic Gradient Descent (SGD) optimizer is used for optimization, with weight decay fixed to 0.0001 and momentum fixed to 0.9.

TABLE 4 GPU SPECIFICATIONS OF THE TRAINING ENVIRONMENT

| Parameter | Specification |
| --- | --- |
| GPU | Nvidia K80 / T4 |
| GPU Memory | 12 GB |
| GPU Memory Clock | 0.82GHz / 1.59GHz |
| Performance | 4.1 TFLOPS / 8.1 TFLOPS |
| No. CPU Cores | 2 |
| RAM | 12 GB |

TABLE 5 CPU SPECIFICATIONS OF THE TRAINING ENVIRONMENT

| Parameter | Specification |
|---|---|
| CPU Model Name | Intel® Xeon® |
| CPU Freq. | 2.30GHZ |
| CPU Family | Haswell |
| No. CPU Cores | 2 |
| RAM | 12 GB |

TABLE 6 CONFIGURATION PARAMETERS OF PROPOSED MODEL

| Parameters | Values |
|---|---|
| BACKBONE | mobilenetv1 |
| IMAGE MAX DIM | 512 |
| IMAGE META SIZE | 15 |
| IMAGE MIN DIM | 800 |
| IMAGE SHAPE | [512 512 3] |
| LEARNING RATE | 0.01 |
| MASK SHAPE | [28,28] |
| RPN_ANCHOR_SCALES | (8, 16, 32, 64, 128) |
| STEPS PER EPOCH | 50 |
| WEIGHT DECAY | 0.0001 |

## VI. Dataset Details

Table 7 gives detailed dataset information. A MISD dataset consists of a total of 918 images which are in JPG G format. The size of an AU image and multiple spliced image is $384 \times 256$ in pixels.

TABLE 7 MISD DATASET DETAILS.

| | |
|---|---|
| Total number of images | 918 ,Authentic = 618,Spliced = 300 |
| Image Format | JPG |
| Image Size | Authentic Image = $384 \times 256$ , Spliced Image= $384 \times 256$ |
| Image Quality | 200-300 dpi |
| Softwares used during MSID construction | • Figma<br>• Background Remover (https://www.remove.bg/) ,<br>• VGG Annotation Tool,<br>• Python script for generating ground truth mask |
| Minimum and Maximum Number of images used for Multiple Image Splicing | Minimum – 3  and  Maximum - 7 |

## VII. Results and Discussion

This section specifies the results of the proposed model for multiple image splicing forgery detection. Tables 8 and 9 show the F1-Score, Precision, and Recall for ResNet (ResNet 51, ResNet 101, ResNet 151) and MobileNet V1. From tables, it is observed that the F1-Score of the proposed model i.e Mask R-CNN model with MobileNet V1 as backbone network outperforms the ResNet models with less number of parameters. Figure 9 shows F1-score, Precision, Recall using Mask RCNN with backbone networks such as MobileNet V1 and variants of ResNet (ResNet 51,101 and 151) for detection of multiple image splicing on our MISD dataset. The model with F1 scores Precision and Recall is represented on the X-axis, while the assessed metrics are represented on the Y-axis.

TABLE 8 F1-SCORE, PRECISION AND RECALL FOR MULTIPLE IMAGE SPLICING DETECTION WITH BACKBONE AS RESNET WITH ITS VARIANTS

| Type of Forgery | Dataset | ResNet 51 | | | ResNet 101 | | | ResNet 151 | | |
| --- | --- | --- | --- | --- | --- | --- | --- | --- | --- | --- |
| | | *F1-Score* | *Precision* | *Recall* | *F1-Score* | *Precision* | *Recall* | *F1-Score* | *Precision* | *Recall* |
| **Multiple Image Splicing** | **MISD** | 0.53 | 0.66 | 0.74 | 0.55 | 0.63 | 0.77 | 0.54 | 0.69 | 0.73 |

TABLE 9 F1-SCORE, PRECISION, AND RECALL FOR MULTIPLE IMAGE SPLICING DETECTION WITH BACKBONE AS MOBILENET V1.0

| Type of Forgery | Dataset | MobileNetV1 | | |
| --- | --- | --- | --- | --- |
| | | F1-Score | Precision | Recall |
| **Multiple Image Splicing** | MISD | 0.67 | 0.73 | 0.62 |

TABLE 10 AVERAGE PRECISION RESULTS ON MULTIPLE IMAGE SPLICING DETECTION WITH BACKBONE AS RESNET WITH ITS VARIANTS

| Type of Forgery | Dataset | ResNet 51 | | | ResNet 101 | | | ResNet 151 | | |
| --- | --- | --- | --- | --- | --- | --- | --- | --- | --- | --- |
| | | Avg. Precision | Avg. Precion0.5 (AP0.5) | Avg. Precion 0.75 (AP0.75) | Avg. Precision | Avg. Precion0.5 (AP0.5) | Avg. Precion0.75 (AP0.75) | Avg. Precision | Avg. Precion0.5 (AP0.5) | Avg. Precion0.75 (AP0.75) |
| **Multiple Image Splicing** | MISD | 0.86 | 0.80 | 0.75 | 0.83 | 0.80 | 0.73 | 0.78 | 0.82 | 0.70 |

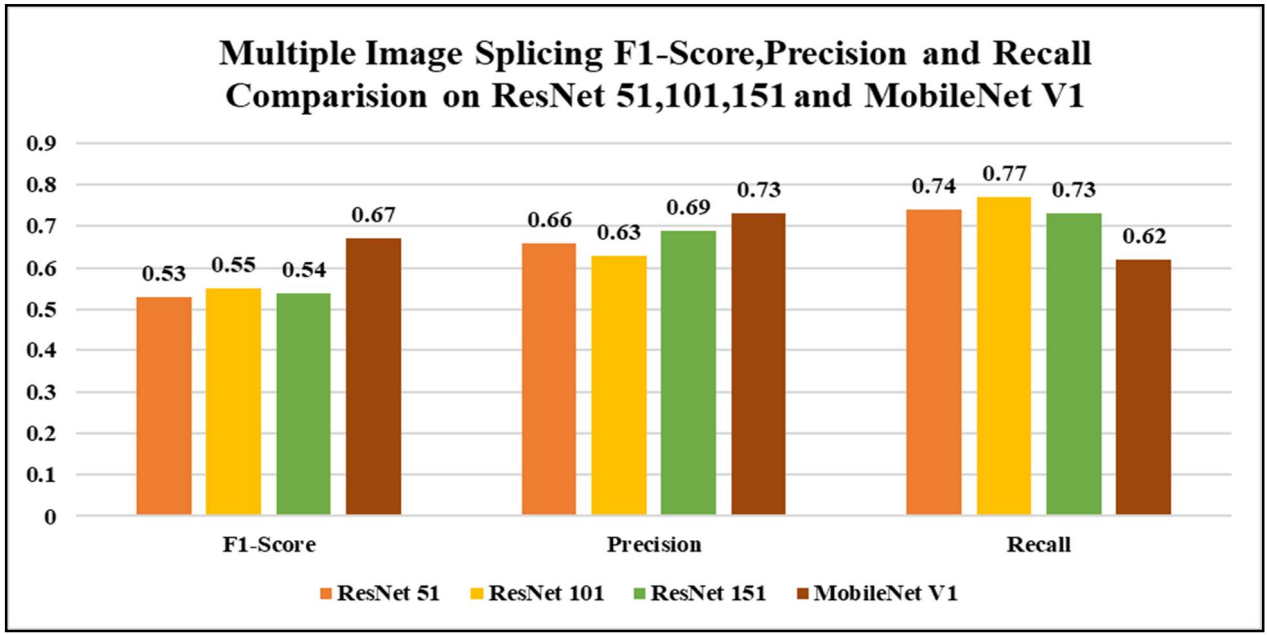

Figure 9. **Multiple Image Splicing F1-Score, Precision and Recall Comparison on ResNet 51,101,151 and MobileNet V1.0**

Tables 10 and 11 show the AP, $AP_{0.5}$, and $AP_{0.75}$ for ResNet (ResNet 51, ResNet 101, ResNet 151) and MobileNet V1 over MISD dataset for $AP_{0.5}$, and $AP_{0.75}$, IOU is 0.5 and 0.75 respectively. Figure 10 shows AP, $AP_{0.5}$, and $AP_{0.75}$ for multiple image splicing forgery detection on our MISD with backbone networks as ResNet (ResNet 51,101,151) and MobileNet V1. Here, the X-axis depicts the model with various average precision values, and Y-axis depicts evaluated metrics.

TABLE 11 AVERAGE PRECISION RESULTS ON MULTIPLE IMAGE SPLICING DETECTION WITH BACKBONE AS MOBILENET V1

| Type of Forgery | Dataset | MobileNetV1 | | |
| --- | --- | --- | --- | --- |
| | | Avg. Precision | Avg. Precion0.5 (AP0.5) | Avg. Precion0.75 (AP0.75) |
| **Multiple Image Splicing** | MISD | **0.85** | **0.88** | **0.73** |

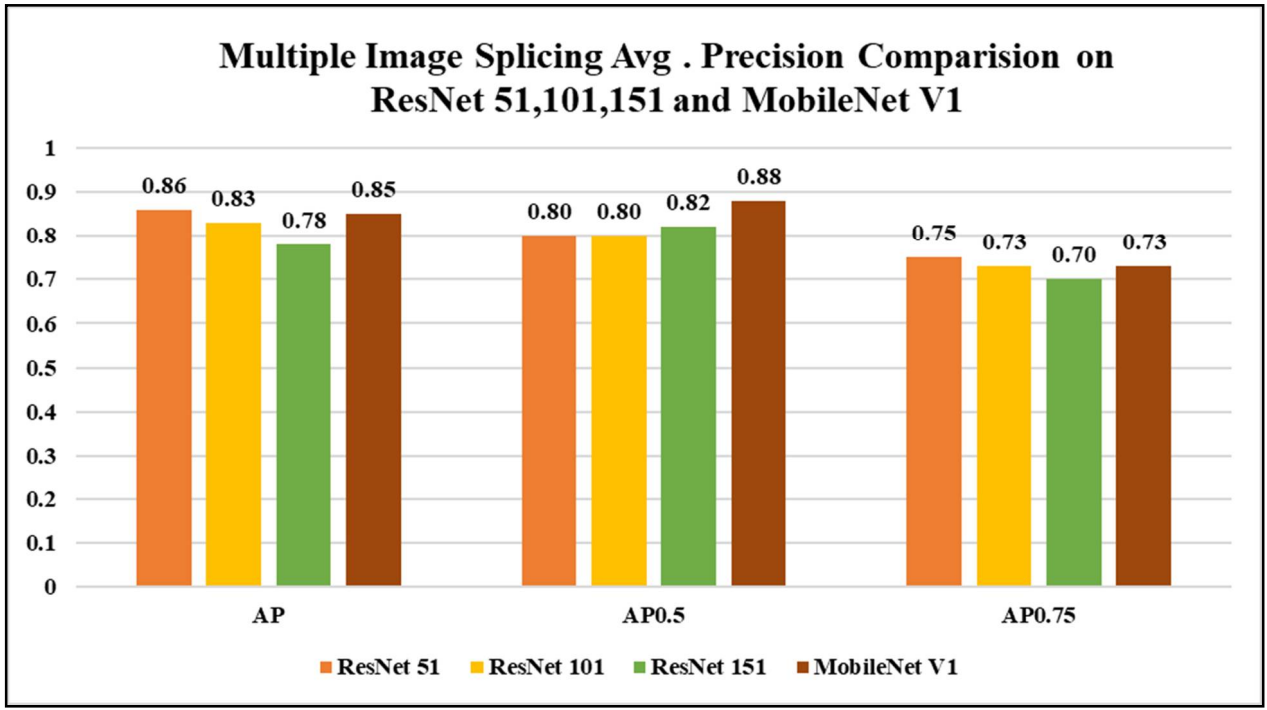

Figure 10. **Multiple Image Splicing Average Precision Comparison on ResNet 51,101,151 and MobileNet V1.0**

After the proposed model produced relatively good results, we used K-fold cross-validation to avoid over-fitting the proposed model to get an indication of the true performance of the model. The proposed model was trained using the k-fold cross validation to evaluate the efficiency of the model. The value of K chosen was 5. The training dataset was divided into 5 subsets or folds randomly, and in each step one of the subsets was used as the validation set and the other 4 folds were used as the training set. The performance of the model is the average metric score obtained over the 5 times training. The average metric scores for the 5-fold cross validation evaluation are shown in the table. During the 5-fold validation, as every image sample gets an opportunity to be a testing sample, unlike randomly picking up the training and testing data, it provided results comparable to the testing results. Tables 12 and 13 show the F1-Score, Precision, and Recall for ResNet (ResNet 51, ResNet 101, ResNet 151) and MobileNet V1 with k-fold cross validation. From tables, it is observed that the F1-Score of the proposed model i.e Mask R-CNN model with MobileNet V1 as backbone network outperforms the ResNet models with less number of parameters. Figure 11 shows F1-score, Precision, Recall using Mask RCNN with backbone networks such as MobileNet V1 and variants of ResNet (ResNet 51,101 and 151) for detection of multiple image splicing on our MISD dataset. The model with F1 scores Precision and Recall is represented on the X-axis, while the assessed metrics are represented on the Y-axis.

TABLE 12 F1-SCORE, PRECISION AND RECALL FOR MULTIPLE IMAGE SPLICING DETECTION WITH BACKBONE AS RESNET WITH ITS VARIANTS WITH K-FOLD CROSS VALIDATION

| Type of Forgery | Dataset | ResNet 50 | | | ResNet 101 | | | ResNet 152 | | |
|---|---|---|---|---|---|---|---|---|---|---|
| | | F1-score | Precision | Recall | F1-score | Precision | Recall | F1-score | Precision | Recall |
| Multiple Image Splicing | MSID | 0.55 | 0.66 | 0.70 | 0.55 | 0.61 | 0.70 | 0.51 | 0.62 | 0.70 |

TABLE 13 F1-SCORE, PRECISION, AND RECALL FOR MULTIPLE IMAGE SPLICING DETECTION WITH BACKBONE AS MOBILENET V1 WITH K-FOLD CROSS VALIDATION

| Type of Forgery | Dataset | MobileNetV1 | | |
|---|---|---|---|---|
| | | F1-Score | Precision | Recall |
| **Multiple Image Splicing** | MISD | 0.68 | 0.75 | 0.62 |

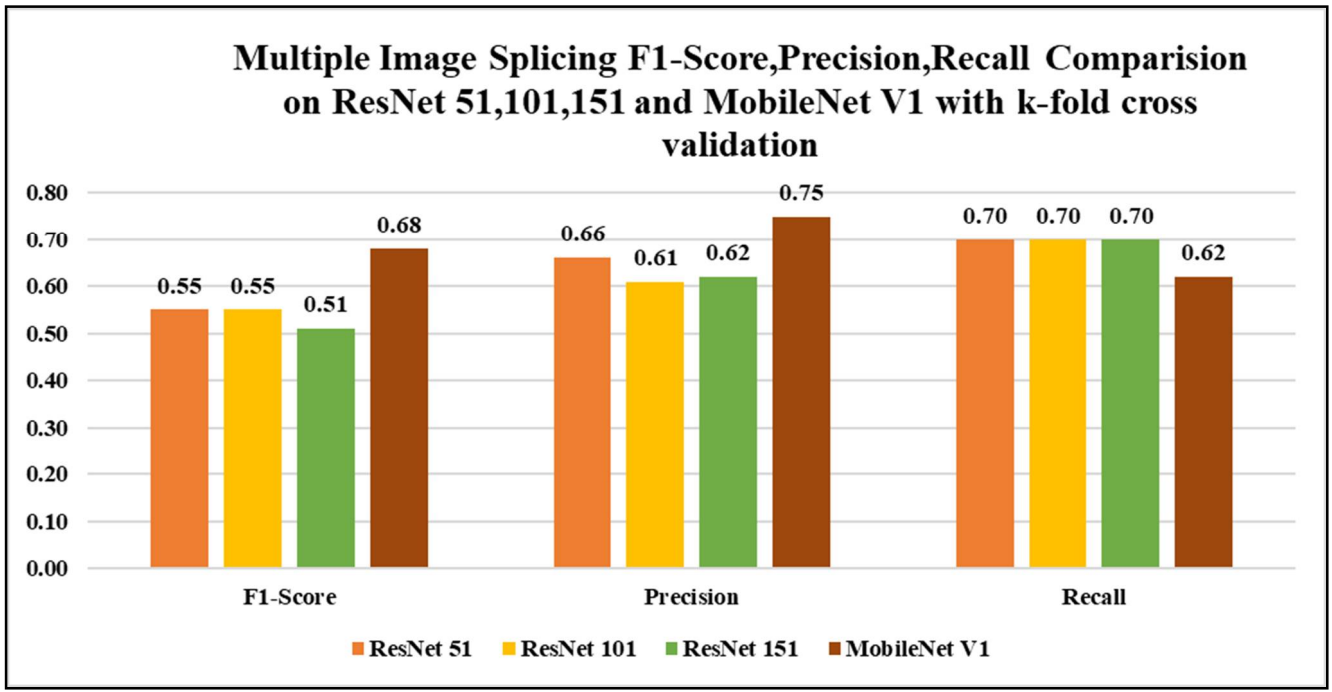

Figure 11. **Multiple Image Splicing F1-Score, Precision and Recall Comparison on ResNet 51,101,151 and MobileNet V1.0 with k-fold cross validation**

Tables 14 and 15 show the AP, $AP_{0.5}$, and $AP_{0.75}$ for ResNet (ResNet 51, ResNet 101, ResNet 151) and MobileNet V1 over MISD dataset using k-fold cross validation for $AP_{0.5}$, and $AP_{0.75}$, IOU is 0.5 and 0.75 respectively. Figure 12 shows AP, $AP_{0.5}$, and $AP_{0.75}$ for multiple image splicing forgery detection on our MISD with backbone networks as ResNet (ResNet 51,101,151) and MobileNet V1 using k-fold cross validation. Here, the X-axis depicts the model with various average precision values, and Y-axis depicts evaluated metrics.

TABLE 14 AVERAGE PRECISION RESULTS ON MULTIPLE IMAGE SPLICING DETECTION WITH BACKBONE AS RESNET WITH ITS VARIANTS WITH k-FOLD CROSS VALIDATION

| Type of Forgery | Dataset | ResNet 51 | | | ResNet 101 | | | ResNet 151 | | |
|---|---|---|---|---|---|---|---|---|---|---|
| | | Avg. Precision | Avg. Precion0.5 (AP0.5) | Avg. Precion 0.75 (AP0.75) | Avg. Precision | Avg. Precion0.5 (AP0.5) | Avg. Precion0.75 (AP0.75) | Avg. Precision | Avg. Precion0.5 (AP0.5) | Avg. Precion0.75 (AP0.75) |
| Multiple Image Splicing | MISD | 0.63 | 0.76 | 0.60 | 0.63 | 0.80 | 0.78 | 0.68 | 0.80 | 0.66 |

TABLE 15 AVERAGE PRECISION RESULTS ON MULTIPLE IMAGE SPLICING DETECTION WITH BACKBONE AS MOBILENET V1.0 VARIANTS WITH k-FOLD CROSS VALIDATION

| Type of Forgery | Dataset | MobileNetV1 | | |
|---|---|---|---|---|
| | | Avg. Precision | Avg. Precion0.5 (AP0.5) | Avg. Precion0.75 (AP0.75) |
| Multiple Image Splicing | MISD | 0.60 | 0.87 | 0.77 |

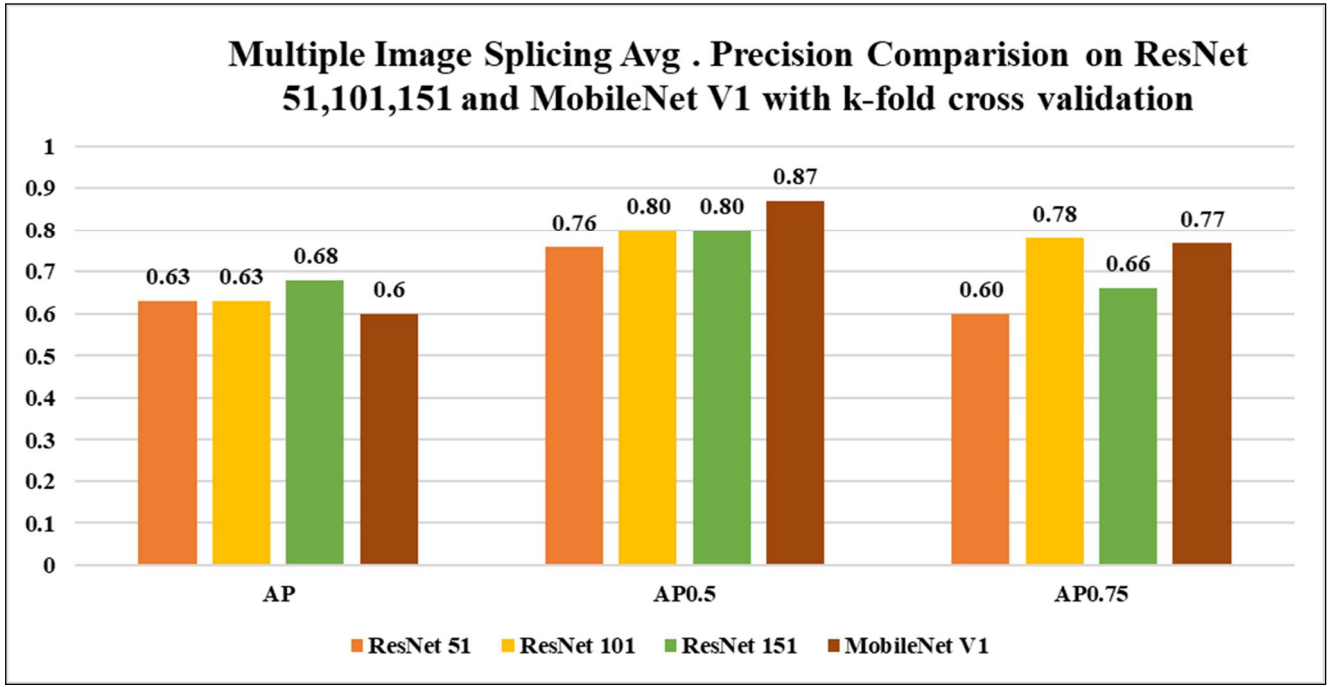

Figure 12. **Multiple Image Splicing Average Precision Comparison on ResNet 51,101,151 and MobileNet V1.0 with k-fold cross validation**

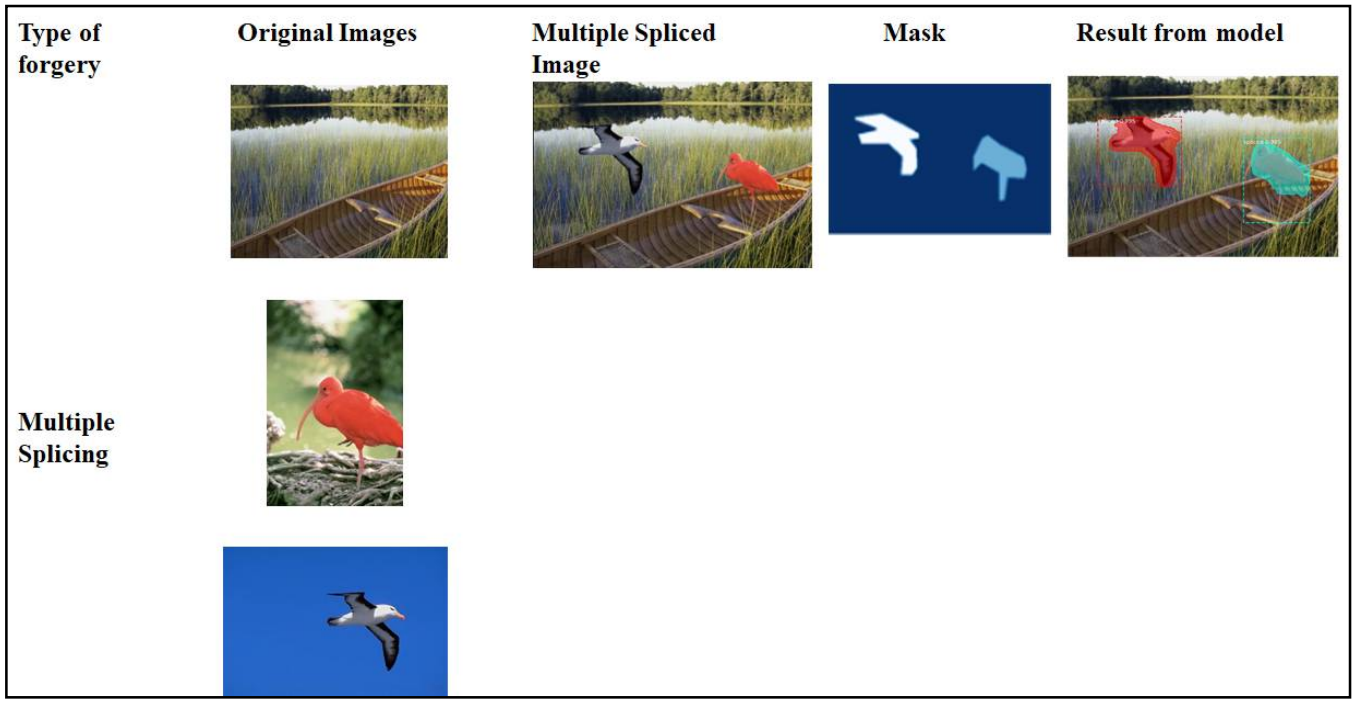

Figure 13. **Multiple Image Splicing – Original Images, Multiple Spliced Image, Mask for a multiple spliced object, and Result from Proposed Model.**

Figure 13 shows the output from the proposed model with original images, multiple spliced image, a mask for the multiple spliced objects. Table 16 shows that the Resnet has a greater number of parameters as compared to MobileNet. MobileNet uses DSC to reduce the model size (number of parameters) and complexity. A network that has many parameters or weights, can provide a better estimate for a large range of functions. The layer of a network stores the parameters or weights in the main memory. So, the fewer the parameters, the faster the model runs. MobileNet offers similar performance as that of Resnet but with a much smaller network due to Depthwise Separable Convolution.

TABLE 16. PARAMETERS, TRAINABLE AND NON-TRAINABLE PARAMETERS WITH BACKBONE NETWORK AS RESNET AND ITS VARIANTS AND MOBILENET V1.0

| Method | Backbone | Parameters | Trainable | Non-Trainable |
|---|---|---|---|---|
| Mask R-CNN | ResNet 51 | 63,733,406 | 63,621,918 | 111,488 |
| | ResNet 101 | 63,733,406 | 63,621,918 | 111,488 |
| | ResNet 151 | 79,446,174 | 79,288,606 | 157,568 |
| | **MobileNet V1** | **23,812,574** | **23,784,542** | **28,032** |

## VIII. Limitations

- *Size of MISD Dataset*
  The size of the dataset is an important factor in determining the performance of the deep learning model. The proposed model training is heavily dependent upon the images with various post-processing operations performed on them. A limited number of images is one of the challenge of this research work.
- *Annotation of data*
  Image annotation is playing an important role in deep learning and machine learning models for image classification segmentation and object recognition. Manual annotation of forged images is reliant on the annotator's knowledge of the data labeling task.

## IX. Conclusion

This research work presents Mask R-CNN with MobileNet V1 as a lightweight model for the detection of multiple image splicing forgery. It also provides a forged percentage score for multiple spliced images. We evaluated the proposed model on our MISD dataset. We have done a comparative analysis of the proposed model with variants of ResNet such as ResNet 51,101, and 151. The proposed model achieves an average precision of 82% on our Multiple Splicing Image Dataset. The configuration of the proposed model is more efficient in terms of computing than variants of ResNet. The evaluation of the proposed model compared to variants of the ResNet network shows that the proposed approach efficiently balanced efficiency and computational costs.

**Abbreviations**
DL- Deep Learning
CV - Computer Vision
CNN - Convolutional Neural Network
FCN - Fully Convolutional Network
SVM - Support Vector Machine
RPN - Region Proposal Network
ROIs - Regions of Interest
Mask R-CNN - Mask Regional Convolutional Neural Network
DSC - DepthZwise Separable Convolution
DSCLs- DepthZwise Separable Convolution Layers
bbox - bounding box
NMS - Nonmax suppression
IOU Intersection Over Union
DWT- Discrete wavelet transform
LBP - Local Binary Pattern
CT- Contourlet Transform
HHT- Hilbert-Huang Transform
DCT - Discrete Cosine Transform
RP- region proposal
ILSVRC- ImageNet Large Scale Visual Recognition Challenge
AU- authentic
SP- Spliced
spatial rich model - SRM